\documentclass[10pt,twocolumn,letterpaper]{article}

\usepackage[pagenumbers]{cvpr}      %

\usepackage{cuted} %
\usepackage{caption} %

\usepackage{microtype}

\usepackage{multirow}
\usepackage{adjustbox}
\usepackage{pifont} %
\usepackage{colortbl} %
\newcommand{\ours}{PCRG}
\newcommand{\tkeep}{\textsc{keep}}
\newcommand{\tcache}{\textsc{cache}}
\newcommand{\tprune}{\textsc{prune}}
\newcommand{\x}{\ensuremath{\times}}

\definecolor{cvprblue}{rgb}{0.21,0.49,0.74}
\usepackage[pagebackref,breaklinks,colorlinks,allcolors=cvprblue]{hyperref}

\title{\ours{}: Training-Free Pruning and Caching of Reference Tokens for Video Generation}
    
\author{Siyuan Wang\quad Haoran Li\quad Yi Yue\quad Tak L. Fung\quad Quan Chen\\
Anorak Tech\\
{\small\url{https://anorak-games.github.io/pcrg/}}}

\begin{document}
\maketitle
\begin{strip}
\centering
\includegraphics[width=\linewidth]{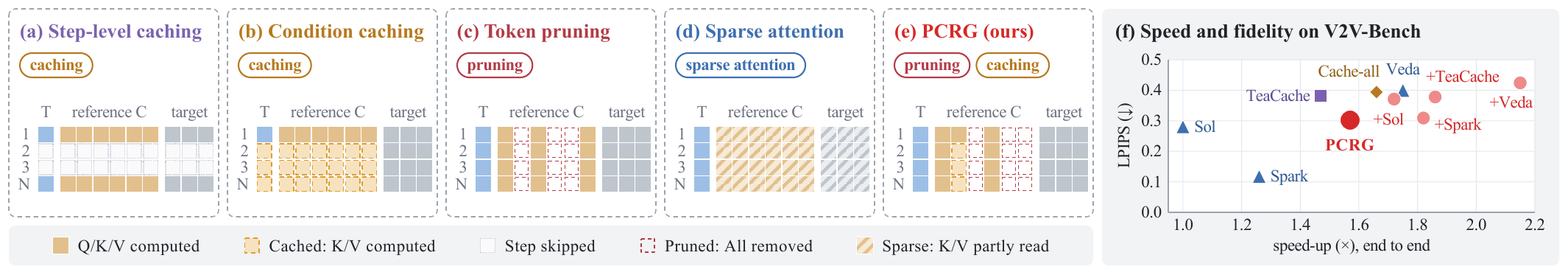}
\captionsetup{hypcap=false}
\captionof{figure}{\textbf{Caching and pruning.} Each grid shows tokens by steps. \textbf{(a)}~Step-level caching skips whole steps. 
\textbf{(b)}~Condition caching keeps all keys and values but stops updating. 
\textbf{(c)} Pruning removes the tokens entirely. 
\textbf{(d)}~Sparse attention keeps every token but reads only part of the keys and values. 
\textbf{(e)} \ours{} combines caching and pruning for the reference token.
\textbf{(f)}~Comparison results on speed and fidelity on V2V-Bench.}

\captionsetup{hypcap=true}
\label{fig:teaser}
\end{strip}
\begin{abstract}
    Reference-to-video diffusion transformers pack the variational autoencoder (VAE) latents of all references and the generated tokens into one full-attention sequence, in which reference tokens make up about half.
Because attention cost grows quadratically with sequence length, these tokens take up an even larger share of the attention computation, yet most of them are static, smooth or irrelevant to the output.
To accelerate reference-conditioned generation, we propose \textbf{pruning} and \textbf{caching} of \textbf{reference} tokens for video \textbf{generation} (\textbf{PCRG}), a training-free, plug-and-play method that uses two signals the model already computes to assign each reference token to one of three tiers, namely pruned, cached or kept.
The first is a low-level \emph{redundancy} $r$ from VAE latents, since a token whose latent barely changes over time or across its neighbors adds little new content.
The second is a high-level \emph{importance} $g$ from the vision-language-model (VLM) conditioner, where the attention that the prompt and any first or last frames pay to a token tells whether the output needs to read it.
Tokens that are not redundant are kept, and redundant tokens are cached if important and pruned otherwise.
Pruned tokens leave the sequence before the first step with all rotary position embedding (RoPE) positions unchanged. Cached tokens have their keys and values computed once and reused, and kept tokens are processed as usual.
On MiniMax-H3, already distilled to 8 timesteps and quantized to int8, PCRG is 1.57$\times$ faster end to end on our V2V-Bench and 1.49$\times$ faster on the public VACE-Benchmark, while the quality of VBench drops by only 0.24\% and 0.17\%, respectively.
PCRG shortens the sequence rather than changing how attention is computed or which steps are run, so it also stacks on step caching and sparse attention. On V2V-Bench it reaches 1.86$\times$ with TeaCache, 1.82$\times$ with Spark at nearly unchanged fidelity and 2.15$\times$ with Veda.
\end{abstract}
\section{Introduction}
\label{sec:intro}
Reference-to-video (R2V) models generate a video that follows reference images and videos, such as segmentation maps or rendered previews, often together with first and last frames (FL-frames) and the prompt. Current R2V models are diffusion transformers (DiTs)~\cite{peebles2023dit,wan2025,chen2026skyreelsv4,yang2026omnivideo2} that condition in context. The variational autoencoder (VAE) latents of every reference join the generated tokens in one full-attention sequence~\cite{jiang2025vace,ju2025fulldit,wang2026refalign,cong2026omnixref}. 
Reference tokens are never generated, yet they pass through every layer at every denoising step, and every reference image or second of reference video lengthens the sequence.

Much of this cost buys little. Tokens of static or smooth regions are redundant, since they add little new content and need not be recomputed at every step. The prompt, moreover, often names only a few objects, so many tokens are irrelevant to the output. A redundant token that is still relevant can therefore have its keys and values computed once and reused - caching, an irrelevant one can be removed from the sequence - pruning, and only the remaining tokens need to be processed as usual.

Training-free accelerators of video DiTs do not exploit this. Step-level caching~\cite{zhao2025pab,liu2025teacache,son2026relational,lv2026navicache,shen2026gpcache} (\cref{fig:teaser}a) treats reference and generated tokens alike. Condition caching (\cref{fig:teaser}b) needs training~\cite{tan2025ominicontrol2,he2025fulldit2,liu2026anchorcache}. Without it, caching every condition token (\emph{cache-all}, as in KV-Edit~\cite{zhu2025kvedit}) keeps all keys and values and freezes the tokens that carry detail. Token reduction (\cref{fig:teaser}c) either restores a dense grid of generated tokens after each block~\cite{bolya2023tomesd,lee2026dito,lee2026orbis}, needs training~\cite{he2025fulldit2,parihar2026keep,he2026refroute}, or covers only image editing~\cite{lin2026topi} and the unmasked context of video edits~\cite{liu2026hetcache}. Sparse attention~\cite{xi2025svg,tan2026adacluster,luo2026inputstable,hu2026dfsattn,han2026veda}, including that of the Sol engine~\cite{nvidia2026sol} (\cref{fig:teaser}d), only thins the attention over an unchanged sequence.

We present \textbf{pruning} and \textbf{caching} of \textbf{reference} tokens for video \textbf{generation} (\ours{}), a training-free, plug-and-play method that assigns each reference token to one of three tiers using two signals the model already computes (\cref{fig:teaser}e and \cref{fig:method}). The low-level \emph{redundancy} $r$ is read from the VAE latents and is high for tokens that are static over time or smooth in space, which add little content. The high-level \emph{importance} $g$ is read from the attention that the prompt and any FL-frames pay to a reference token in the single causal pass of the vision-language model (VLM) conditioner~\cite{bai2025qwen3vl}, and tells whether the output needs to read that token. Before the first denoising step, tokens that are not redundant are kept (\tkeep{}), and redundant ones are cached (\tcache{}) if important, with their keys and values computed once and reused, or otherwise pruned (\tprune{}) from the sequence. The remaining tokens keep their original rotary positions. Both signals are read outside the DiT, so \ours{} never needs the DiT's attention maps and leaves its attention kernels, dense or sparse, unchanged.

On MiniMax-H3, already distilled to 8 timesteps and quantized to int8, \ours{} runs 1.57\x{} faster end to end on our V2V-Bench, which covers structure-to-video, editing and multi-reference requests, and 1.49\x{} faster on the public VACE-Benchmark, while VBench quality drops by only 0.24\% and 0.17\%, respectively (\cref{fig:teaser}f).
Because it shortens the sequence rather than changing which steps run or how attention is computed, it also stacks on step caching and sparse attention. On V2V-Bench, it reaches 1.86\x{} with TeaCache~\cite{liu2025teacache}, 1.82\x{} with Spark at almost the same fidelity and 2.15\x{} with Veda~\cite{han2026veda} (\cref{sec:accel}).

In summary, our main contributions are:
\begin{itemize}
\item We show that in R2V generation with in-context conditioning, reference tokens make up about half of the sequence and an even larger share of the attention computation, yet most of them are static, smooth or irrelevant to the output.
\item We propose \ours{}, which cuts the cost of redundant reference tokens in two ways: it prunes those the output does not need from the sequence and caches the keys and values of those it does need. 
Each token is assigned by low-level redundancy from the VAE latents and high-level importance from the attention in the VLM.
\item On V2V-Bench and VACE-Benchmark, \ours{} gives speed-ups of 1.57\x{} and 1.49\x{} with almost unchanged quality on top of step distillation and quantization.
To our knowledge, it is the first training-free method that both prunes and caches the reference tokens of R2V generation.
\end{itemize}
\section{Related Work}\label{sec:related}
\paragraph{Reference-to-video generation.} Most current video generators are diffusion transformers~\cite{peebles2023dit} on 3D-VAE latents~\cite{kong2024hunyuanvideo,wan2025,chen2026skyreelsv4}. Conditions were first injected through cross-attention or an added control branch~\cite{zhang2023controlnet}. Reference-to-video and subject-consistent generation models~\cite{liu2025phantom,wang2026refalign,wang2026anyid} now mostly condition in context instead: the latents of the reference images join the generated tokens in one full-attention sequence~\cite{jiang2025vace,wang2026refalign}, so that every generated token can read every reference token. FullDiT~\cite{ju2025fulldit} also appends structure videos such as depth or pose maps, and unified editing models treat the source video as one more reference~\cite{cong2026omnixref}. Text encoders have become multimodal LLMs~\cite{kong2024hunyuanvideo,yang2026omnivideo2,chen2026skyreelsv4} that read the references together with the prompt and attend to the reference regions the prompt refers to. MiniMax-H3 reads the references, the FL-frames and the prompt in one causal Qwen3-VL pass~\cite{bai2025qwen3vl}, which also yields the importance of \ours{}. The cost of in-context conditioning grows with every reference~\cite{wu2026loomvideo}, and this is the cost \ours{} cuts.

\paragraph{Acceleration and token reduction.} Feature caching reuses block outputs or attention across steps~\cite{zhao2025pab,liu2025teacache,son2026relational,lv2026navicache,shen2026gpcache}, sparse attention exploits the locality of video attention~\cite{xi2025svg,tan2026adacluster,luo2026inputstable,hu2026dfsattn,han2026veda}, and step distillation~\cite{nie2026tmd}, quantization~\cite{zhao2025viditq,li2025svdquant} and full-stack engines~\cite{nvidia2026sol} cut the number and cost of operators (see the survey~\cite{shao2026efficientvdm}). All of them work on a sequence of fixed length, whereas \ours{} shortens it and therefore stacks with them. Our results are measured on an int8 model distilled to 8 timesteps (7 transformer evaluations), and sparse attention on top of \ours{} adds further speed (\cref{sec:accel}). Token reduction in diffusion models merges or matches the \emph{generated} tokens~\cite{bolya2023tomesd,lee2026dito,lee2026orbis}, which must still be restored to a dense grid after each block, whereas reference tokens are never produced and can be removed outright.

\paragraph{Attention-based token pruning.} VLMs prune visual tokens by the attention that text tokens pay to them~\cite{chen2024fastv}, a score that attention sinks distort~\cite{li2026sinkpruner} and that concentrates the budget unless coverage or distribution alignment is enforced~\cite{zhu2026coverpruner,chen2026otprune}. \ours{} reads this attention inside the VLM conditioner to prune the reference tokens of the DiT, normalizes it against sinks, and keeps coverage by selecting redundancy locally (\cref{sec:method}).

\paragraph{Efficient conditioning.} Most methods train the model for it. EasyControl~\cite{zhang2025easycontrol} and OminiControl2~\cite{tan2025ominicontrol2} reuse condition keys and values under a restricted attention, and OminiControl2 additionally compresses condition tokens to the most relevant ones.
FullDiT2~\cite{he2025fulldit2} also prunes and caches video in-context conditions: a learned scorer selects reference tokens to shorten the sequence, and their features are cached across steps and layers. 
RefRoute~\cite{he2026refroute} compresses and routes reference images, and LIVEditor~\cite{shao2026liveditor} learns a sparse attention that prunes most in-context tokens of video edits; it can also run without training, but replaces the attention of the DiT. LoomVideo~\cite{wu2026loomvideo} adds the conditions to the generated tokens instead of appending them to the sequence. Without training, KV-Edit~\cite{zhu2025kvedit} reuses the keys and values of an image's unedited background, and diffusion language models approximately reuse those of their prompt~\cite{wu2026fastdllm,liu2026dllmcache}. Applied to an R2V model, this is our \emph{cache-all} baseline (\cref{sec:accel}). 
Closest to ours are training-free methods that reduce references. ToPi~\cite{lin2026topi} prunes image-editing references, but needs the DiT's own attention maps. HetCache~\cite{liu2026hetcache} prunes and caches tokens of video edits, but only in the unmasked context of a masked edit. HiLo-Token~\cite{you2026hilotoken} compresses unedited image regions by spatial frequency, a low-level cue akin to our redundancy, with no signal of what the prompt needs. \ours{} instead targets reference-to-video generation with arbitrary reference videos and images, reads both of its signals outside the DiT so that its attention kernels, dense or sparse, stay in place, and combines low-level redundancy with high-level importance to decide per token between keeping, caching and pruning.
\section{Preliminaries}\label{sec:bg}
\paragraph{Flow matching.} Like most current video DiTs~\cite{wan2025}, MiniMax-H3 is trained by flow matching~\cite{lipman2023flow,liu2023rectified}, in which a clean latent $x_0$ and noise $\epsilon\sim\mathcal N(0,I)$ are joined by the straight path
\begin{equation}
x_t=(1-t)\,x_0+t\,\epsilon,\qquad t\in[0,1],
\label{eq:fm-path}
\end{equation}
and the transformer $v_\theta$ regresses its velocity given the conditions $c$,
\begin{equation}
\mathcal L(\theta)=\mathbb E_{x_0,\epsilon,t}\big\|v_\theta(x_t,t,c)-(\epsilon-x_0)\big\|_2^2 .
\label{eq:fm-loss}
\end{equation}
Sampling integrates from $t_0=1$ to $t_N=0$ in $N$ Euler steps, each a full transformer evaluation,
\begin{equation}
x_{k+1}=x_k+(t_{k+1}-t_k)\,v_\theta(x_k,t_k,c),
\label{eq:fm-euler}
\end{equation}
where $x_k$ is the current latent, $v_\theta(x_k,t_k,c)$ is the velocity that the transformer predicts from $x_k$, the time $t_k$ and the conditions $c$, namely the direction in which to move, and $t_{k+1}-t_k$ is the step size.

\paragraph{In-context reference-to-video generation.} 
MiniMax-H3 runs full attention over one packed sequence $z_k=[\,T \mid C \mid K \mid x_k\,]$, where $T$ is the VLM output, $C$ the VAE latents of the references, $K$ the VAE latents of the FL-frames, and $x_k$ the noisy latents of the target video and audio at step $k$, and only $x_k$ is denoised. 
Phantom~\cite{liu2025phantom} and VACE~\cite{jiang2025vace} pack reference images in the same way.

\paragraph{The VLM.} The conditioner of MiniMax-H3 is Qwen3-VL-32B~\cite{bai2025qwen3vl}.
It reads all conditions as one token sequence without a chat template, namely each reference after its label, then the FL-frames as further pictures, then the prompt as plain text.
A video is sampled at 2\,fps and read in two-frame blocks, each preceded by its timestamp.
The ViT cuts each image or block into $16{\times}16$-pixel patches and merges $2{\times}2$ neighbouring patches into one LLM token.
Since the LLM attends causally, reference tokens never see the FL-frames or the prompt, while the FL-frames and the prompt attend to the references.

\paragraph{Caching and pruning.} Training-free accelerators either reuse computation across the steps of \cref{eq:fm-euler} or remove tokens. Step-level caching~\cite{zhao2025pab,liu2025teacache} reuses the output of whole steps or blocks when the input changes little. It treats all tokens alike, and after distillation few steps are left to skip. Condition caching~\cite{tan2025ominicontrol2,zhang2025easycontrol,liu2026anchorcache} and token pruning~\cite{lin2026topi,liu2026hetcache} act on tokens instead. In every layer $\ell$,
\begin{equation}
\begin{aligned}
\text{cache } (k>1):\ & \mathbf K^{(\ell)}_{C,k}\approx\mathbf K^{(\ell)}_{C,1},\ \ \mathbf V^{(\ell)}_{C,k}\approx\mathbf V^{(\ell)}_{C,1},\\
\text{prune } (\text{all } k):\ & z_k\leftarrow z_k[S],\ \ p\leftarrow p[S],
\end{aligned}
\label{eq:cache-prune}
\end{equation}
where $S$ is the set of kept tokens and $p$ are their rotary positions, which are not renumbered. Caching no longer updates $C$, but every query still reads its keys and values. Pruning saves both but also loses the pruned content. \ours{} keeps, caches or prunes each reference token by its redundancy and importance (\cref{fig:teaser}e).

\section{Method}\label{sec:method}
\begin{figure*}[t]\centering
\includegraphics[width=\linewidth]{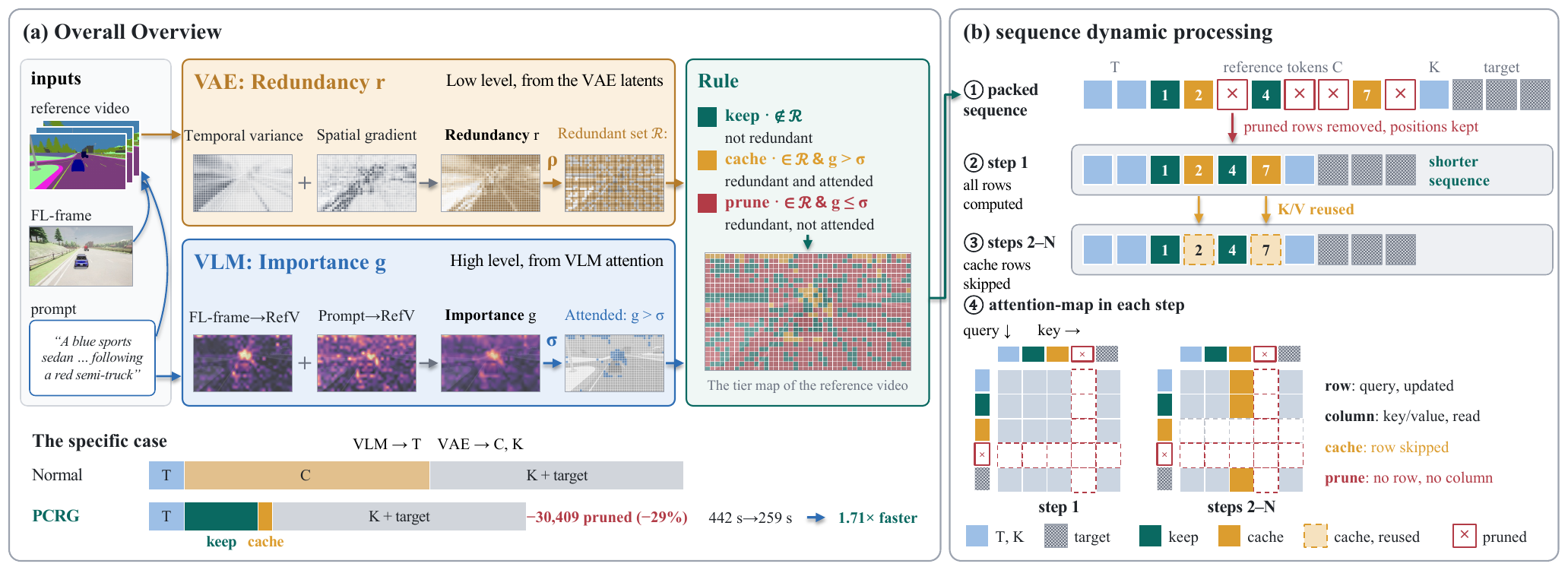}
\caption{\textbf{Overview of \ours{}.} \textbf{(a)}~Before step~1, the redundancy $r$ of the VAE latents selects the redundant set $\mathcal R$, the most redundant fraction $\rho$ of each block. The importance $g$ from the VLM marks the cells with $g>\sigma$ as attended.~\textbf{(b)}~Pruned rows leave the packed sequence, and the remaining tokens keep their RoPE (\protect\ding{172}). Step~1 computes every remaining row (\protect\ding{173}). Steps~2--$N$ skip the cached rows and reuse their keys and values from step~1 (\protect\ding{174}). In the attention maps (\protect\ding{175}), from step~2 on, a cached token keeps its key column but loses its query row, and a pruned token has neither.}

\label{fig:method}
\end{figure*}
A request may carry several references, each an image or a video, and \ours{} treats them one by one, an image as a single latent frame. 
Before the first denoising step, we decides once how each reference token is computed (\cref{fig:method}a). The diffusion transformer then runs on a shorter sequence in which some rows are frozen (\cref{fig:method}b).
A \tkeep{} token is computed at every step, a \tcache{} token only at the first step, and a \tprune{} token never. Later steps still read the keys and values of a \tcache{} token.
Two signals decide the tier map, both read from passes the model already runs.
The low-level \emph{redundancy} $r$ from the VAE latents tells whether a token's content is also carried by its neighbours as shown in~\cref{sec:redundancy}, and a high-level \emph{importance} $g$ from the VLM tells whether the generated tokens need to attend to it as shown in~\cref{sec:xattn}.
Redundancy decides which tokens may leave the per-step computation, and importance decides which of them remain attendable. \Cref{sec:tiers} gives the rule and its execution.
    
\subsection{Low-level redundancy from the VAE}\label{sec:redundancy}
MiniMax-H3, like other video DiTs~\cite{wan2025,kong2024hunyuanvideo}, denoises in the latent space of a video VAE.
Since the VAE is trained to reconstruct pixels, its latents record low-level appearance, such as colour, edges and texture, rather than high-level semantics~\cite{zheng2026rae}.
The references are encoded by the same VAE into the tokens $C$ (\cref{sec:bg}), so each reference token is a patch of this latent.
Two tokens with similar latents therefore look alike, and the latent shows low-level redundancy directly.
The reference token can be excluded from per-step computation if its information is replicated in spatio-temporal neighbourhoods — either the same spatial region across adjacent frames, or adjacent regions within the same frame.
We split each reference latent into the DiT's $2{\times}2$ patches and average each patch into $\bar z_c(\tau,h,w)$, where $c$ is the channel, $\tau$ the latent frame and $(h,w)$ the cell.
Each cell thereby spans all latent frames of the reference.
Temporal redundancy is unique to video inputs, and arises when a cell exhibits minimal variation over time.
We quantify this temporal variation as per-cell variance across \(\tau\), averaged over the \(C_z\) channels of the VAE latent and \(C_z{=}24\) in H3:
\begin{equation}
\mathrm{var}(h,w)=\frac{1}{C_z}\sum_{c=1}^{C_z}\mathrm{Var}_{\tau},\bar z_c(\tau,h,w).
\end{equation}
It first calculates the temporal variance of patch-averaged features along the latent frame dimension for each channel, then takes the mean across channels to obtain a scalar measuring the cell’s temporal stability.
Spatial redundancy arises within a frame when the latent feature is smooth across adjacent cells.
We measure this via the spatial gradient of channel- and frame-averaged latents \(\tilde z(h,w)=\operatorname{mean}_{c,\tau}\bar z_c(\tau,h,w)\), computed as forward differences between adjacent cells:
\begin{equation}
\mathrm{edge}(h,w)=|\partial_h\tilde z|+|\partial_w\tilde z|.
\end{equation}
Both metrics are min–max normalized across all cells within each reference sample:
\begin{equation}
u\leftarrow\frac{u-\min u}{\max u-\min u},\qquad u\in\{\mathrm{var},\mathrm{edge}\},
\end{equation}
and further fused into a unified redundancy score:
\begin{equation}
r=\begin{cases}1-\tfrac12\big(\mathrm{var}+\mathrm{edge}\big), & \text{video},\\ 1-\mathrm{edge}, & \text{image},\end{cases}
\end{equation}
The resulting redundancy $r\in[0,1]$ is higher for cells that are temporally static and spatially smooth, as shown in the VAE row of \cref{fig:method}a.
Because $r$ is computed per cell, all tokens of a cell, one per latent frame, fall into the same tier.

\paragraph{Local selection.} 
Large smooth regions (e.g., roads, sky) appear highly redundant, but wholesale removal of such regions would eliminate reference coverage for the corresponding frame areas.
\ours{} thus adopts a local selection strategy for redundant cells.
Cells are partitioned into non-overlapping $3{\times}3$ blocks, and the highest-$r$ cells in each block are marked as redundant, so that about the same fraction $\rho$ of every block becomes redundant.
Formally, let $q$ denote the relative rank of a cell within its block $B(h,w)$, where rank~1 corresponds to the highest redundancy,
\begin{equation}
q(h,w)=\frac{\operatorname{rank}_{B(h,w)}\,r(h,w)-\frac12}{|B(h,w)|}\in(0,1).
\end{equation}
The redundant set $\mathcal R$ comprises the $\mathrm{round}(\rho\,|\Omega|)$ cells with the smallest $q$ across all $|\Omega|$ reference cells, with ties resolved in favour of larger $r$.
Global ranking across the full frame would eliminate entire smooth regions in a video, causing the output to deviate from the unpruned one.
Compared with a fixed threshold on $r$, the fraction-based strategy marks the same share of every reference as redundant, regardless of the overall smoothness of the reference.
\paragraph{Edits and images.}
For editing tasks, the output aligns with the source video region-wise, requiring full spatial coverage, while redundancy arises primarily across adjacent frames.
\ours{} thus prunes entire latent frames rather than spatial cells.
From the $F$ latent frames of the source video, a fraction $D=0.6$ is discarded, leaving $F'=F-\mathrm{round}(DF)$ evenly spaced frames from the first to the last,
\begin{equation}
\mathcal F=\Big\{\mathrm{round}\Big(\frac{i\,(F-1)}{F'-1}\Big):\ i=0,\dots,F'-1\Big\},
\end{equation}
where $i$ indexes the retained frames, from the first latent frame ($i=0$) to the last ($i=F'-1$).
All tokens of the discarded frames are pruned without caching (\cref{sec:cd}).
For reference images, ranking is applied globally across all cells of the image, and the smoothest 50\% of cells form $\mathcal R$ ($\rho_{\text{img}}=0.5$).

\subsection{High-level importance from the VLM}\label{sec:xattn}
Whether a reference region matters depends on the conditions, such as the prompt and any FL-frames.
The prompt may specify objects to be reproduced or replaced, while FL-frames indicate content that the reference must anchor.
In the VLM's causal sequence as shown in \cref {sec:bg}, reference tokens precede FL-frames and the prompt, so reference-side signals, e.g., raw features or similarity to the prompt, cannot adapt to these downstream conditions, as \cref {sec:cand} shows.
\ours{} therefore reads the attention in the other direction. The prompt and the FL-frames come after the references, so their tokens can attend to every reference cell, and a cell that receives much of this attention is important.
For a downstream condition, let Q denote the queries of its tokens at VLM layer L, and \(A_L(q,j)\) denote the head-averaged softmax attention weight from query q to reference token j.
The total attention received by reference token j from this condition is defined as:
\begin {equation}
a^{L}(j)=\sum_{q\in Q} A_L (q,j).
\end {equation}
For video inputs, the attention map is further averaged across temporal blocks.
The importance score g, computed as the average of the FL-frame attention map at layer m and the prompt attention map at layer n, is then resampled to match the cell grid \((h,w)\):
\begin {equation}
g=\tfrac12\big (a^{m}_{\text{FL}}+a^{n}_{\text{prompt}}\big),
\end{equation}where we set \(m=24\) and \(n=40\) in our implementation.
A cell is considered attended and important when \(g>\sigma\), where \(\sigma\) is the importance threshold.
FL-frame-to-reference attention locates content shared between FL-frames and the reference, while prompt-to-reference attention identifies regions corresponding to the text description.
The prompt map follows the prompt most at layer 40 and the FL-frame map finds the foreground best at layer 24, and both weaken in deeper layers.
This importance calculation is executed within the conditioner's native forward pass: it only accesses query rows and processes one attention head at a time, bringing no extra overhead to conditioning time.

\subsection{Prune, cache or keep}\label{sec:tiers}

\paragraph{Rule.}
Given the redundant set $\mathcal R$ (\cref{sec:redundancy}) and importance score $g$ (\cref{sec:xattn}), we assign each reference cell to one of three tiers:
\begin{equation}
\text{tier}=
\begin{cases}
\tkeep & (h,w)\notin\mathcal R,\\
\tcache & (h,w)\in\mathcal R \text{ and } g>\sigma,\\
\tprune & (h,w)\in\mathcal R \text{ and } g\le\sigma.
\end{cases}
\end{equation}
These assignments form a tier map for each reference (\cref{fig:method}a).
Non-redundant cells retain full computation, important redundant cells remain available to attention, and the rest are removed.
We use $\rho=0.7$ for videos, $\rho_{\text{img}}=0.5$ for images, and $\sigma=0.4$.

\paragraph{Execution.}
Pruned rows are removed from the packed sequence before the first denoising step (\cref{fig:method}b, \ding{172}).
All remaining tokens retain their original rotary position IDs.
We evaluate cached rows in full at the first denoising step (\cref{fig:method}b, \ding{173}) and approximate their states at later steps by reusing the saved input to each layer.
Let $\mathcal M$ denote the cached rows and $\mathcal U$ all other retained rows; for each layer $\ell$ and step $k>1$, we set (\ding{174})
\begin{equation}
h^{(\ell)}_{\mathcal M,k}=h^{(\ell)}_{\mathcal M,1}.
\end{equation}
Attention is computed only for queries from $\mathcal U$:
\begin{equation}
\mathbf O_{\mathcal U}=\operatorname{softmax}\!\Big(
\frac{\mathbf Q_{\mathcal U}\,[\mathbf K_{\mathcal U};\mathbf K_{\mathcal M}]^{\top}}{\sqrt d}
\Big)\,[\mathbf V_{\mathcal U};\mathbf V_{\mathcal M}].
\end{equation}
For $k>1$, cached tokens thus supply keys and values but no queries, whereas pruned tokens supply none (\ding{175}).
Let $n$ denote the number of tokens in the original sequence before pruning.
For each step $k>1$, the attention matrix products and tokenwise feed-forward layers have the following cost ratios relative to full computation on that sequence:
\begin{equation}\label{eq:cost}
c_{\text{attn}}=\frac{|\mathcal U|\,(|\mathcal U|+|\mathcal M|)}{n^2},
\qquad
c_{\text{FFN}}=\frac{|\mathcal U|}{n}.
\end{equation}

\paragraph {Cache size.}
Every redundant token is either pruned or cached: pruning reduces computational cost, while caching preserves fidelity to the unpruned output. The threshold \(\sigma\) controls the trade-off between the two (\cref {sec:cd}).
Existing condition caching methods~\cite {liu2026anchorcache} cache all condition tokens, which only eliminates query-side computation while retaining all key-value pairs in the sequence. In contrast, \ours {} prunes the majority of redundant tokens and caches only a small fraction, achieving higher acceleration (\cref {sec:accel}).
Leveraging the precomputed importance score, our method selects attendable redundant tokens at no additional cost, making the cache allocation adaptive to the prompt.
Tasks requiring high-fidelity reference reproduction, such as video editing, benefit from a larger cache proportion.

\section{Experiments}\label{sec:exp}
\subsection{Setup}
\paragraph{Model.} 
We use MiniMax-H3 ref2va with its distilled sampler of 8 timesteps, running an int8-quantized transformer on a single NVIDIA RTX PRO 6000 GPU.
Speed-ups are measured end to end against the identical pipeline without \ours {}, and are therefore obtained on top of step distillation and quantization.

\paragraph{Benchmarks.} 
Our \emph {V2V-Bench} contains 92 real-world requests, covering instruction-based editing, editing guided by a reference image, and generation conditioned on a reference video; it will be released alongside the paper.
The public \emph {VACE-Benchmark}~\cite {jiang2025vace} provides an additional 240 requests spanning control, inpainting and outpainting, extension, composition, and reference-to-video.

\paragraph{Baselines.} 

All baselines run on the same pipeline.
The first group retains the full token sequence and instead accelerates inference by skipping denoising steps or computing attention sparsely.
It comprises the step-level caching method TeaCache~\cite{liu2025teacache} (\(\delta{=}0.06\)) and three sparse-attention methods: Spark, Veda~\cite{han2026veda}, and the sparse attention of the Sol engine~\cite{nvidia2026sol} (\(\tau{=}1.2\)).
The second group, like \ours {}, operates exclusively on reference tokens.
\emph {cache-all}, the training-free counterpart of condition caching~\cite {zhang2025easycontrol,he2025fulldit2,liu2026anchorcache}, computes all reference rows in the first denoising step and reuses their keys and values in all subsequent steps.
We additionally include two reference-downscaling baselines: resizing the short edge to 384 pixels, or retaining every other frame prior to encoding.

\paragraph{Metrics.} 
Following the convention of training-free accelerators for video DiTs~\cite {zhao2025pab,liu2025teacache,xi2025svg}, we compare each output against the unpruned result of the same seed using PSNR, SSIM, and LPIPS.
We further report absolute generation quality as VBench-Q, the average of six prompt-free VBench dimensions: subject consistency, background consistency, motion smoothness, temporal flickering, aesthetic quality, and imaging quality.
Speed-ups are averaged across requests, with each request timed against an unpruned baseline run in the same session.
We additionally adopt the pairwise VLM judge from SolarRenderer~\cite {fang2026solarrenderer}, replacing its default Gemini with Qwen3-VL-32B~\cite {bai2025qwen3vl}.

\subsection{Main results}\label{sec:v2vb}\label{sec:accel}

\begin{figure*}[t]\centering
\includegraphics[width=\linewidth]{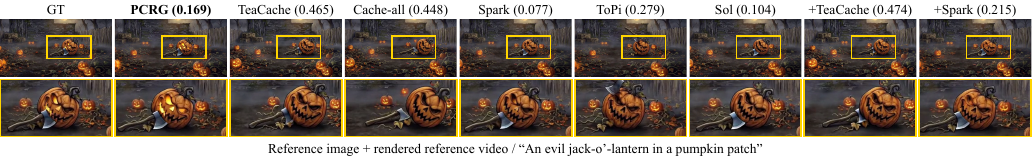}
\caption{
Qualitative comparison on a V2V-Bench request. GT is the unpruned output, and the number in parentheses is the LPIPS to it. +TeaCache and +Spark combine \ours{} with TeaCache and Spark. 
}
\label{fig:cmp}
\end{figure*}

On our V2V-Bench, requests are generated at their native resolution and duration, with reference tokens accounting for roughly 20–60\% of the total sequence.
The larger the proportion of reference tokens, the greater the speedup delivered by \ours {}.
\Cref {tab:bench-accel} compares training-free accelerators on both benchmarks under identical settings.
On V2V-Bench, with all baselines timed in the same session, \ours{} achieves a 1.57\(\times\) speedup while reducing VBench-Q by only 0.20 points (0.24\%).
It outperforms TeaCache, cache-all, and Veda across fidelity metrics.
Spark achieves the highest fidelity but only reaches a 1.26\(\times\) speedup.
Moreover, \ours {} is orthogonal to these approaches and can be stacked on top of them to further improve inference speed without compromising generation quality.
\Cref{tab:bench-accel} reports that on the public VACE-Benchmark, \ours{} achieves a 1.49\(\times\) speedup, lowering VBench-Q by only 0.15 points (0.17\%).
At a resolution of 832$\times$480, VACE-Benchmark sequences are roughly half the length of V2V-Bench sequences, so the gains from stacking are more modest.
When combined with TeaCache, \ours{} reaches a 1.75\(\times\) speedup; with Spark, it reaches 1.62\(\times\).
The pairwise VLM judge cannot distinguish \ours {} from the unpruned output~(GT) on either benchmark.
\Cref{fig:cmp} shows an example: \ours{} keeps the subject of the unpruned output, while TeaCache, cache-all and ToPi change it.
Notably, a positive net score is not a benefit for an accelerator: it indicates that the output deviates from the unpruned result in a direction the judge happens to prefer.

\begin{table}[t]
\centering
\footnotesize
\setlength{\tabcolsep}{3.5pt}
\caption{Accelerators on V2V-Bench and VACE-Benchmark. Speed, PSNR, SSIM, and LPIPS is end to end against the ground truth. Judge is the pairwise VLM score against the GT.
VB-Q is the mean of six VBench dimensions in percent; changes quoted in the text are computed before rounding.
The Bold and UNDERLINE denote the best and second-best results, respectively.}
\label{tab:bench-accel}
\begin{tabular}{l cccccc}
\toprule
Method & Speed & PSNR$\uparrow$ & SSIM$\uparrow$ & LPIPS$\downarrow$ & Judge$\uparrow$ & VB-Q \\ %
\midrule
\multicolumn{7}{c}{\textit{V2V-Bench}} \\
Unpruned & 1.00$\times$ & -- & -- & -- & -- & 84.23 \\
TeaCache~\cite{liu2025teacache} & 1.47$\times$ & 16.64 & 0.569 & 0.381 & \underline{$+.18$} & 84.44 \\ %
Spark & 1.26$\times$ & \textbf{23.46} & \textbf{0.769} & \textbf{0.118} & $-.04$ & 84.22 \\ %
Cache-all & 1.66$\times$ & 16.25 & 0.562 & 0.394 & $.00$ & 84.10 \\ %
ToPi~\cite{lin2026topi} & 1.59$\times$ & 17.55 & 0.595 & 0.321 & $-.07$ & 84.15 \\
Veda~\cite{han2026veda} & 1.75$\times$ & 17.00 & 0.577 & 0.400 & $\mathbf{+.31}$ & \textbf{85.33} \\ %
Sol~\cite{nvidia2026sol} & 1.00$\times$ & \underline{18.50} & \underline{0.620} & \underline{0.281} & -- & 84.18 \\
\rowcolor{gray!12} \textbf{\ours{}} & 1.57$\times$ & 17.89 & 0.605 & 0.302 & $-.08$ & 84.03 \\
\rowcolor{gray!12} \quad +TeaCache & \underline{1.86}$\times$ & 16.56 & 0.567 & 0.378 & $+.07$ & 84.14 \\ %
\rowcolor{gray!12} \quad +Spark & 1.82$\times$ & 17.77 & 0.601 & 0.309 & $-.15$ & 84.01 \\
\rowcolor{gray!12} \quad +Veda & \textbf{2.15}$\times$ & 16.51 & 0.568 & 0.424 & $+.12$ & \underline{85.28} \\ %
\rowcolor{gray!12} \quad +Sol & 1.72$\times$ & 16.54 & 0.571 & 0.371 & -- & 84.13 \\
\midrule
\multicolumn{7}{c}{\textit{VACE-Benchmark}} \\
Unpruned & 1.00$\times$ & -- & -- & -- & -- & 85.20 \\
TeaCache & 1.40$\times$ & \underline{18.32} & 0.596 & 0.297 & \underline{$+.02$} & \underline{85.27} \\ %
Spark & 1.18$\times$ & \textbf{23.11} & \textbf{0.755} & \textbf{0.118} & $-.02$ & 85.26 \\ %
Cache-all & 1.54$\times$ & 16.50 & 0.563 & 0.350 & $\mathbf{+.04}$ & \textbf{85.35} \\ %
Veda & 1.33$\times$ & 17.84 & 0.602 & 0.286 & \underline{$+.02$} & 85.07 \\ %
\rowcolor{gray!12} \textbf{\ours{}} & 1.49$\times$ & 17.73 & \underline{0.604} & \underline{0.276} & $-.03$ & 85.06 \\ %
\rowcolor{gray!12} \quad +TeaCache & \textbf{1.75}$\times$ & 17.13 & 0.580 & 0.312 & $-.06$ & 85.10 \\ %
\rowcolor{gray!12} \quad +Spark & \underline{1.62}$\times$ & 17.53 & 0.598 & 0.285 & $-.06$ & 85.11 \\ %
\rowcolor{gray!12} \quad +Veda & \underline{1.62}$\times$ & 16.49 & 0.567 & 0.347 & $-.08$ & 84.81 \\ %
\bottomrule
\end{tabular}
\end{table}

\begin{table}[t]
\centering
\small
\setlength{\tabcolsep}{4pt}
\caption{Redundant-set selection on VACE-Benchmark with a reference video: by $r$, at random, or by inverted $r$. The Bold and UNDERLINE denote the best and second-best results, respectively.}
\label{tab:rsel}
\begin{tabular}{l cccc}
\toprule
Selection & Speed & PSNR$\uparrow$ & SSIM$\uparrow$ & LPIPS$\downarrow$ \\
\midrule
\multicolumn{5}{l}{\emph{Prune only, 60\% of the cells}} \\
by $r$ & 1.57$\times$ & \textbf{17.89} & \textbf{0.613} & \textbf{0.291} \\
random & 1.56$\times$ & \underline{17.54} & \underline{0.605} & \underline{0.307} \\
inverted $r$ & 1.56$\times$ & 17.28 & 0.595 & 0.315 \\
\midrule
\multicolumn{5}{l}{\emph{Cache and prune, 70\% of the cells}} \\
\rowcolor{gray!12} \textbf{by $r$ (\ours{})} & 1.65$\times$ & \textbf{17.31} & \textbf{0.598} & \textbf{0.317} \\
random & 1.65$\times$ & \underline{17.20} & \underline{0.594} & \underline{0.324} \\
inverted $r$ & 1.65$\times$ & 17.00 & 0.590 & 0.330 \\
\bottomrule
\end{tabular}
\end{table}

\subsection{Ablation studies}\label{sec:ablation}
\begin{figure}[t]\centering
\includegraphics[width=\linewidth]{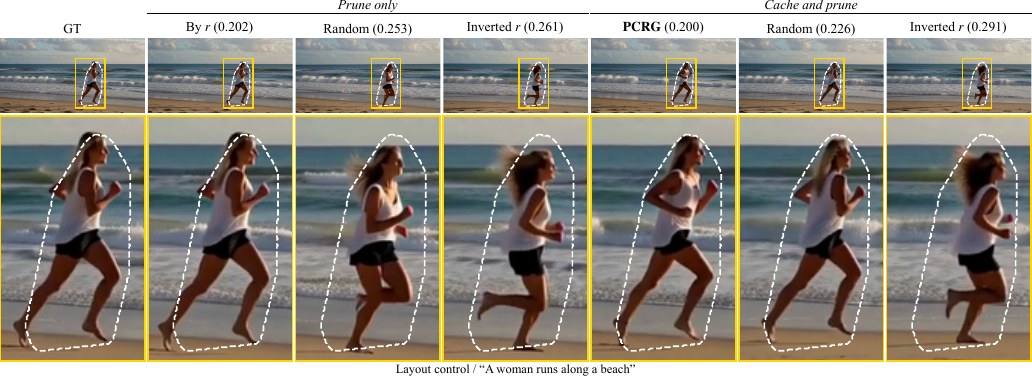}
\caption{
Redundant-set selection on a layout-control request of VACE-Benchmark 
The white dashed line is the runner outline from the layout control input, drawn at the same place in every panel. The number in parentheses is the LPIPS to GT. 
}
\label{fig:rabl}
\end{figure}

\paragraph{Low-level redundancy.} 
To verify the effectiveness of the redundancy score r derived from VAE latents, we fix the size of the redundant set and vary only its selection strategy: by r, at random, or by inverted r, on VACE-Benchmark requests with a reference video (\cref {tab:rsel}).
When 60\% of the cells are pruned without caching, selection by r yields the output closest to the unpruned result, and significantly outperforms both random selection and inverted r.
When 70\% of cells are marked and then either cached or pruned following \ours{}, selection by r remains the best and random selection second, yet the margin narrows, because under all three strategies the important cells are cached rather than pruned.
Both settings consistently show that r effectively captures latent redundancy and distinguishes redundant cells from informative ones.
\Cref{fig:rabl} shows an example on a layout-control request. As we can see, the \ours{} and by r can effectively preserve GT consistency.

\paragraph{High-level importance.}\label{sec:cand} 

\begin{figure}[t]\centering
\includegraphics[width=\linewidth]{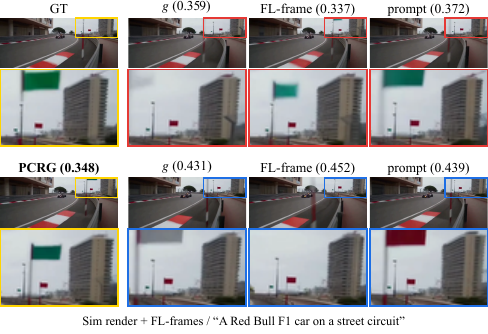}
\caption{
Pruning 60\% of the reference cells by importance, no cache. 
Red boxes means the least attended first dropped. 
Blue boxes means the most attended first dropped. 
yellow boxes is GT and \ours{}.
}
\label{fig:gabl}
\end{figure}

To verify that the importance score g extracted from the VLM identifies which reference tokens matter, we ablate its two signals—FL-frame\(\to\)reference and prompt\(\to\)reference attention—on the V2V-Bench requests with FL-frames.
We prune a fixed 60\% of reference-video cells without caching, so all variants run at the same speed; cells are selected by g or by either signal, pruning the least- or most-attended cells first (\cref {tab:gabl}).
Across all variants, pruning the most-attended cells substantially degrades fidelity, confirming that the output relies on both the prompt and FL-frame.
The FL-frame signal carries more weight than the prompt signal. 
When the most-attended cells are pruned, FL-frame-only reaches an LPIPS of 0.372, worse than prompt-only at 0.348 and worse than g at 0.367, removing what the FL-frame attends to is more damaging than removing what the prompt attends to.
\Cref {fig:gabl} illustrates that pruning the most-attended cells introduces content absent from GT, whereas pruning the least-attended cells stays close to it.
Thus importance identifies cells whose removal is costly, while redundancy is the better criterion for which cells can be removed (\cref{tab:rsel}). 
\ours{} therefore uses r to form the redundant set and g only to keep its attended cells in cache.

\begin{table}[t]
\centering
\small
\setlength{\tabcolsep}{4pt}
\caption{Pruning by importance on the V2V-Bench with an FL-frame. 
The Bold and UNDERLINE denote the best and second-best results, respectively.}
\label{tab:gabl}
\begin{tabular}{l ccc}
\toprule
Pruned cells & PSNR$\uparrow$ & SSIM$\uparrow$ & LPIPS$\downarrow$ \\
\midrule
\multicolumn{4}{l}{\emph{Least attended first}} \\
by $g$ & \textbf{18.18} & \textbf{0.621} & \underline{0.317} \\
by FL-frame$\to$ref.\ only & \underline{18.04} & 0.617 & \textbf{0.314} \\
by prompt$\to$ref.\ only & 18.00 & \underline{0.618} & 0.324 \\
\midrule
\multicolumn{4}{l}{\emph{Most attended first}} \\
by $g$ & \underline{17.25} & \underline{0.594} & \underline{0.367} \\
by FL-frame$\to$ref.\ only & 17.20 & 0.593 & 0.372 \\
by prompt$\to$ref.\ only & \textbf{17.68} & \textbf{0.605} & \textbf{0.348} \\
\midrule
\rowcolor{gray!12} \textbf{\ours{}} & 18.19 & 0.636 & 0.310 \\
\bottomrule
\end{tabular}
\end{table}

\paragraph{Prune and cache.}\label{sec:cd} 

We fix the redundant set of \ours{}, selected within \(3{\times}3\) blocks by \(r\) with \(\rho{=}0.7\), and vary only the threshold \(\sigma\) that splits it between the cache and the pruned tier, on VACE-Benchmark requests with a reference video (\cref{tab:cd}).
\begin{table}[t]\centering\small\setlength{\tabcolsep}{4pt}
\caption{
Splitting the redundant set of \ours{} between cache and prune tiers by $\sigma$, on VACE-Benchmark with a reference video. 
}
\label{tab:cd}
\begin{tabular}{lcccc}
\toprule
Split & Speed & PSNR$\uparrow$ & SSIM$\uparrow$ & LPIPS$\downarrow$ \\
\midrule
Cache all redundant & 1.27\x & 18.11 & 0.618 & 0.281 \\
$\sigma{=}0.2$ & 1.48\x & 17.63 & 0.607 & 0.304 \\
\rowcolor{gray!12} \textbf{$\sigma{=}0.4$ (\ours{})} & 1.65\x & 17.31 & 0.598 & 0.317 \\
$\sigma{=}0.6$ & 1.69\x & 17.20 & 0.593 & 0.323 \\
$\sigma{=}0.8$ & 1.70\x & 17.26 & 0.594 & 0.321 \\
prune all redundant & 1.71\x & 17.19 & 0.593 & 0.324 \\
\bottomrule
\end{tabular}
\end{table}

Caching every redundant cell stays closest to the GT~(unpruned output) but gains only 1.27\(\times\), and pruning every one is the fastest at 1.71\(\times\) but the farthest from it.
As \(\sigma\) increases, more redundant cells are pruned instead of cached, and the speed and the distance from GT grow together.
At \(\sigma{=}0.4\), \ours{} reaches 1.65\(\times\), over 90\% of the speedup of pruning every redundant cell, and stays about as close to GT.

\paragraph{Reference images.}\label{sec:refimg}
The same redundancy criterion also prunes reference images, spatially.
On the V2V-Bench requests with reference images, we compare \ours{} with variants that leave the images untouched, prune only the images, or prune or cache every redundant image cell, as shown in~\cref{tab:imgabl}.
Images are a small part of the sequence, so pruning them alone gains only 1.06\x{} and keeps the output close to GT.
Added to video pruning, image pruning raises the speed-up from 1.43\x{} to 1.56\x{} and is faster on every request.
\begin{table}[t]
\centering
\small
\setlength{\tabcolsep}{4pt}
\caption{
Pruning reference images on the V2V-Bench requests with reference images. 
\emph{Images only} and \emph{video only} apply \ours{} to one kind of reference, and the last two rows prune or cache every redundant image cell instead of splitting them by $\sigma$. 
The Bold and UNDERLINE denote the best and second-best results, respectively.
}
\label{tab:imgabl}
\begin{tabular}{l ccccc}
\toprule
Configuration & Speed & PSNR$\uparrow$ & SSIM$\uparrow$ & LPIPS$\downarrow$ & DINO$\uparrow$ \\
\midrule
Unpruned & 1.00$\times$ & -- & -- & -- & \textbf{0.614} \\
\midrule
Images only & 1.06$\times$ & \textbf{21.21} & \textbf{0.686} & \textbf{0.190} & \underline{0.610} \\
Video only & 1.43$\times$ & \underline{18.43} & \underline{0.615} & \underline{0.290} & 0.605 \\
\rowcolor{gray!12} \textbf{\ours{}} & \textbf{1.56$\times$} & 18.09 & 0.607 & 0.307 & 0.605 \\
\midrule
\multicolumn{6}{l}{\emph{Every redundant image cell}} \\
Pruned & \textbf{1.56$\times$} & 18.07 & 0.604 & 0.308 & 0.607 \\
Cached & \underline{1.49$\times$} & 18.18 & 0.609 & 0.301 & 0.609 \\
\bottomrule
\end{tabular}
\end{table}

\section{Conclusion}\label{sec:conclusion}

In in-context reference-to-video generation, reference tokens make up about half of the sequence and an even larger share of the attention computation, yet most of them are static, smooth or irrelevant to the output.
Without training, \ours{} sorts them into three tiers by two signals the model already computes, the low-level redundancy of the VAE latents and the high-level importance from the VLM conditioner.
Tokens that are not redundant are kept, and redundant ones are cached if important and pruned otherwise.
On MiniMax-H3, already distilled and quantized, \ours{} is 1.57\x{} faster end to end on V2V-Bench and 1.49\x{} faster on VACE-Benchmark.
It also stacks on step caching and sparse attention.
{
    \small
    \bibliographystyle{ieeenat_fullname}
    \bibliography{refs}
}

\end{document}